\documentclass[letterpaper, 10 pt, conference]{ieeeconf}

\IEEEoverridecommandlockouts                              

\usepackage[utf8]{inputenc}
\usepackage{comment}
\usepackage{booktabs}

\usepackage{graphicx}
\usepackage{caption}
\usepackage{subcaption}
\usepackage{amsmath}
\usepackage{verbatim}
\usepackage{amsfonts}
\usepackage{multirow}
\usepackage{changepage}
\usepackage{tabularx}
\usepackage{adjustbox}
\usepackage{mathtools}
\usepackage{amssymb}
\usepackage{soul}
\usepackage{color}
\usepackage[noend]{algpseudocode}

\usepackage{color}
\usepackage{tikz}
\usepackage{flushend}
\usetikzlibrary{shapes.geometric, arrows}
\usepackage[linesnumbered,ruled]{algorithm2e}
\usepackage{multirow}
\usepackage{soul,color}
\usepackage{graphicx}
\usepackage[colorlinks]{hyperref}
\usepackage{fancyhdr}
\usepackage{comment}
\usepackage{epsfig}
\usepackage{bm}
\usepackage{float}
\usepackage{cuted}
\allowdisplaybreaks

\makeatletter
\def\algbackskip{\hskip-\ALG@thistlm}
\makeatother

\newcommand\oprocendsymbol{\hbox{$\square$}}
\newcommand\oprocend{\relax\ifmmode\else\unskip\hfill\fi\oprocendsymbol}

\usepackage{cite}

\newcommand\blue[1]{{\color{blue} #1}}

\newtheorem{theorem}{Theorem}

\newtheorem{remark}{Remark}

\renewcommand{\theenumi}{(\roman{enumi}}

\makeatletter
\let\NAT@parse\undefined
\makeatother

\renewcommand{\baselinestretch}{0.94}

\title{\LARGE \bf
Gaussian Lane Keeping: A Robust Prediction Baseline
}

\author{
 David Isele\textsuperscript{1*}  \hspace{1.2cm}  
Piyush Gupta\textsuperscript{1*}  \hspace{1.2cm}  Xinyi Liu\textsuperscript{1,2}
\hspace{1.2cm}
Sangjae Bae\textsuperscript{1} 
\thanks{
\textsuperscript{1} Honda Research Institute, San Jose, CA, 95134, USA. \texttt{\{disele, piyush\_gupta, sbae\}@honda-ri.com } \
\textsuperscript{2} University of Michigan, Ann Arbor, MI, 48104, USA. \texttt{xinyiww@umich.edu}}
\thanks{\textsuperscript{*} indicates equal contribution}
\thanks{This work has been supported by Honda Research Institute, USA.}
}

\begin{document}

\maketitle
\thispagestyle{plain}
\pagestyle{plain}
\pagenumbering{gobble}
\begin{abstract}
Predicting agents' behavior for vehicles and pedestrians is challenging due to a myriad of factors including the uncertainty attached to different intentions, inter-agent interactions, traffic (environment) rules, individual inclinations, and agent dynamics. Consequently, a plethora of neural network-driven prediction models have been introduced in the literature to encompass these intricacies to accurately predict the agent behavior. Nevertheless, many of these approaches falter when confronted with scenarios beyond their training datasets, and lack interpretability, raising concerns about their suitability for real-world applications such as autonomous driving. Moreover, these models frequently demand additional training, substantial computational resources, or specific input features necessitating extensive implementation endeavors. In response, we propose Gaussian Lane Keeping (GLK), a robust prediction method for autonomous vehicles that can provide a solid baseline for comparison when developing new algorithms and a sanity check for real-world deployment. We provide several extensions to the GLK model, evaluate it on the CitySim dataset, and show that it outperforms  the neural-network based predictions.


\end{abstract}


\section{Introduction}
Trajectory prediction is a heavily researched topic with numerous applications including Autonomous Driving (AD). Recent efforts have focused on multi-modal and interactive prediction models \cite{cheng2023gatraj}, often utilizing deep learning to handle complex interdependencies \cite{varadarajan2022multipath++}. 
Although deep learning models exhibit high performance on static datasets, they often struggle with transferring this performance to out-of-distribution datasets and real-world applications~\cite{9991836}. Observations from researchers suggest that a constant velocity prediction model often provides a more robust baseline in such scenarios~\cite{scholler2019simpler}. Moreover, many researchers and practitioners prefer to employ more reliable and computationally efficient methods for their systems~\cite{dirckx2023optimal}.


Our goal is to develop robust baseline trajectory prediction models, aiming to enhance the existing constant velocity baseline. By robustness, we mean that the resulting trajectories are reliable and interpretable across a wide range of scenarios. When failures do occur, they should be consistent and predictable, thereby providing a strong and dependable prediction baseline. Prioritizing both robustness and ease of implementation, we introduce prediction methods that demonstrate significant improvements over constant velocity, address specific challenges highlighted in the literature such as multi-modality and inter-agent interactions, and have proven exceptionally reliable in real-world autonomous driving testing. 
Note that our focus is not to claim state-of-the-art performance (although we do show that these baselines are competitive), but to provide reliable and interpretable models that can aid in establishing baseline behavior and debugging. By adopting these baselines as comparisons, researchers can gain clarity on the expected potential improvements, aiding the decision-making process regarding the necessity of employing more advanced methods for their specific application. Furthermore, once a researcher decides to invest effort into integrating, training, and converting a new prediction algorithm into a robotics middleware such as a ROS node~\cite{quigley2009ros} for real-world usage, these baselines serve as valuable tools for verifying and debugging performance effectively.


The prediction methods for AD can be classified into three approaches \cite{karle2022scenario}: physics-based, pattern-based, and planning-based. Physics-based methods leverage kinematic principles based on the current state, and they typically prove highly effective for short-term predictions. 
Furthermore, the knowledge of kinematics facilitates high computational efficiency. The examples include the constant velocity model and Intelligent Driver Model (IDM) \cite{idm-def}. Pattern-based methods, on the other hand, rely on datasets to identify and categorize potential patterns using situational information. Machine learning techniques, such as Support Vector Machines (SVM)~\cite{cortes1995support} and Hidden Markov Models (HMM)~\cite{aoude2012driver}, are commonly employed for pattern recognition, which is then mapped to predefined sets of trajectories. 
Planning-based methods estimate the trajectory of each agent by assuming that they are guided by an optimal or non-optimal planner \cite{tian2021anytime}. Model Predictive Control (MPC) \cite{mayne2011tube}, deep learning methods (e.g., imitation learning \cite{torabi2018behavioral}, and reinforcement learning \cite{bouton2020reinforcement}) are often utilized for their capability of obtaining an optimal trajectory (or human-like driving behaviors). The pattern-based and planning-based methods adopt a holistic approach, considering overall behaviors inferred from the dataset for predictions rather than focusing solely on immediate instances. However, they are typically tailored to specific scenarios, as generalizing them necessitates high-quality datasets and extensive training and tuning efforts. Furthermore, the computational demands associated with these methods can become bottlenecks when applied to multiple agents simultaneously. In contrast, physics-based methods serve as excellent baseline techniques because of their (i) exceptional computational efficiency, making them highly scalable for multiple agents, (ii) reduced reliance on extensive datasets, and (iii) practical implementability, as they are analytical models that can be readily applied.

We present a set of physics-based prediction models that serve as robust prediction baselines for most applications and can aid in the development and verification of new algorithms when they are not sufficient. Starting from constant velocity prediction as a foundation model, we consider various extensions that include lane snapping, probabilistic lane keeping, curvature, multiple intentions, and interactivity, the latter of which have been the focus of recent works \cite{10160890,cui2019multimodal,shi2024mtr++}.
We assess our baseline models using the CitySim dataset and present simulation results. We examine the limitations of various commonly used evaluation metrics and offer tools for enhanced visualization and comparison of different prediction algorithms' performance.


There are three major contributions to the literature. (i) We propose a set of robust trajectory prediction baselines that are reliable, easy to implement, independent of a dataset (training-free), and highly interpretable. (ii) We provide a comparative analysis of baselines on the CitySim dataset~\cite{zhang2023citysim} that can be a solid reference for cases where a simpler solution can outperform highly involved solutions such as a neural-network-based prediction. 
\noindent (iii) We provide tools, such as sorted error plots, to improve the understanding of algorithm capabilities and to visually compare the strengths and weaknesses of different prediction algorithms.



\section{Background}\label{sec:background}

In this section, we explore the strengths and limitations of the constant velocity prediction model as a baseline and introduce alternative models aimed at addressing its shortcomings.

\subsection{Constant Velocity (CV) prediction as a baseline} 
Physics-based prediction models, such as Constant Velocity (CV) prediction, offer many advantages. They do not rely on environmental assumptions, require no training, are often expressible in closed form, and can be quickly implemented.
The CV prediction model is a widely used baseline in the literature \cite{bae2020cooperation, chen2018continuous} due to these characteristics, making it a preferred choice among researchers. Moreover, modeling Newton's first law of motion yields predictions that remain robust across a wide range of scenarios. Even in instances where the constant velocity assumption is invalidated, such as when a vehicle is braking, the estimation might still accurately capture the vehicle heading. Considering that agents typically adhere to Newtonian motion for a significant duration, CV predictions often serve as a reliable baseline and generally yield commendable performance on average. Particularly in specific contexts like highway driving, they demonstrate notable competitiveness \cite{moradipari2022predicting}.

\subsection{Lane Snapping with Constant Velocity (LS-CV)}

CV prediction's primary limitation arises in handling turns. Consequently, on datasets emphasizing intersections, it is often easier to outperform CV predictions. Even in highway scenarios, the subtle curvature of the road frequently causes the prediction to deviate from the lane, especially when considering long prediction horizons. However, by integrating lane information with constant velocity, a technique we refer to as lane snapping\footnote{Lane snapping can also be viewed as constant velocity in Frenet–Serret coordinates.}, this limitation can be addressed. Incorporating lane data and predicting constant velocity along the lane restores CV as a highly competitive baseline. This approach can also be extended to pedestrian scenarios by broadening the concept of lanes to encompass crosswalks, sidewalks, or the shortest paths to predicted goal locations. While deep learning methods can outperform lane snapping with constant velocity (LS-CV) on a number of datasets, they are often challenged when tested on road geometries different from the training data. We illustrate this in Section \ref{Sec:dl}.


\subsection{Intelligent Driver Model (IDM)}\label{subsec:IDM} 
A major limitation of employing constant velocity with lane snapping is its inability to account for variations in velocity. To address this, it might be intuitive to consider modeling constant acceleration. However, this approach is not ideal due to two reasons.
\begin{figure}
    \centering
    \includegraphics[trim={0 0 0 1cm},clip,width=0.9\columnwidth]{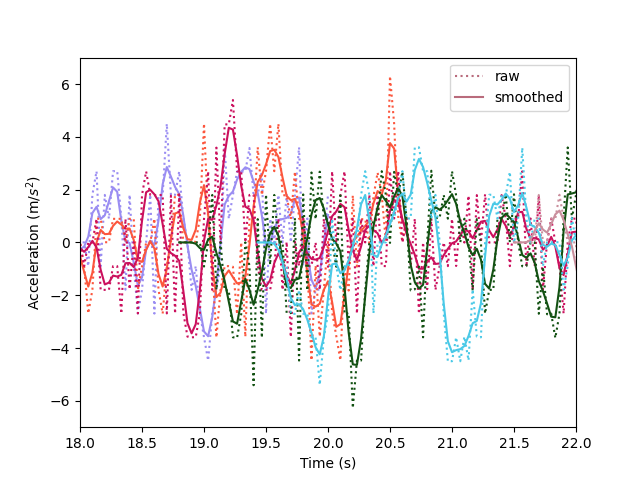}
    \caption{Accelerations of various traffic agents from an arbitrary time window in the CitySim dataset. Each color represents a different agent. It's evident that the accelerations are noisy and do not exhibit a constant profile.}
    \label{fig:acc}
\end{figure}

The first reason is the challenge of dealing with acceleration, which for other agents, is often derived from velocity or position measurements, leading to noisy data. 
Figure~\ref{fig:acc} illustrates this issue, displaying accelerations derived by differentiating the speed of various traffic agents from a random instance in the CitySim Dataset \cite{zhang2023citysim}.
The second reason is that, apart from instances of zero acceleration, constant accelerations are not commonly observed in practice. This can be observed in Fig.~\ref{fig:acc}, where oscillations around zero are prevalent, and these fluctuations are not well approximated by constants.

Therefore, instead of modeling constant acceleration, an alternative approach to account for velocity variations is to utilize a physics-based interactive model such as the Intelligent Driver Model (IDM)~\cite{idm-def} to actively predict velocity or acceleration at each time step. IDM is a widely used mathematical model in traffic engineering that elucidates the behaviors of individual drivers within various traffic scenarios. It simulates and predicts how drivers adjust their speed to maintain safe following distances, considering variables such as desired speed, vehicle spacing, and reaction times. The IDM model calculates acceleration ($a$) as follows:
\begin{align}\label{eq:IDM_a}
    a &= a_{\max}\left[ 1 - \left( \frac{v}{v_0} \right)^{4} -  \left( \frac{s^*}{s} \right)^{2}  \right],
\end{align}
where, $s^*$ represents the desired safety gap (minimum following distance), defined as:
\begin{align}\label{eq:IDM_s}
s^* &= s_0 + s_1 \sqrt{\frac{v}{v_0}} +  v.T+ \frac{v(v - v_{lead})}{2\sqrt{a_{\max}b}}.
\end{align}
 The parameters $v_0$, $s_0$, $s_1$, $T$, $a_{\max}$, $b$ pertain to the IDM model, signifying the driver's desired speed, preferred minimal gap, quantitative agreement parameter, driver's desired time headway, maximum comfortable acceleration, and comfortable deceleration, respectively.


\section{Gaussian Lane Keeping}\label{Sec:GLK_iterative}

Although CV prediction proves reliable for short duration by considering the vehicle's heading, its efficacy diminishes over longer durations due to its failure to incorporate lane information and variations in the vehicle's trajectory. Conversely, while employing LS-CV improves the modeling of changes in heading compared to CV prediction, it introduces a new challenge: excessive confidence in an agent's adherence to the lane trajectory. This poses problems when the agent deviates from the lane, such as entering an unmapped area like a parking lot or a driveway. Additionally, it may overestimate the precision with which the agent follows the lane center. This becomes particularly concerning when cars navigate close to the lane boundary, whether due to distraction or intentional avoidance of obstacles on the other side. It is essential to recognize that while modeling these deviations is important, they are generally either small lateral deviations or infrequent occurrences. Consequently, they have minimal impact on metrics evaluating the entire dataset, a point that will be reiterated throughout this paper.


To address this, we propose a Gaussian Lane Keeping (GLK) model designed to combine the advantages of both CV prediction and LS-CV models. Specifically, GLK operates as a probabilistic model that prioritizes CV prediction (with constant heading) for short-term duration, while gradually leaning towards following the current lane trajectory over longer periods. 




The modeling of GLK utilizes Bayesian statistics 
to combine multiple probabilistic prediction models. Let $X_{t-1} = [p_{x, t-1}, p_{y, t-1}, v_{x, t-1}, v_{y, t-1}]$ denote the state vector consisting of the $x$-$y$ components of the position and velocity of the vehicle at time $t-1$. We model the constant velocity prediction as: 
\begin{equation}\label{eq:cv_model_wrong}
    X_t^{cv} = \begin{bmatrix}
        1 & 0 & \Delta t & 0 \\ 0 & 1 & 0 & \Delta t \\ 0 & 0 & 1 & 0\\ 0 & 0 & 0 & 1
    \end{bmatrix}X_{t-1} + \begin{bmatrix}
        \epsilon_{p_x}^{cv} \\ \epsilon_{p_y}^{cv} \\ \epsilon_{v_x}^{cv}  \\ \epsilon_{v_y}^{cv}
    \end{bmatrix} := AX_{t-1} + \epsilon^{cv}
\end{equation}
where $\Delta t$ is the time step and $\epsilon_{p_x}^{cv}, \epsilon_{p_y}^{cv}, \epsilon_{v_x}^{cv}, \epsilon_{v_y}^{cv}$ are the zero mean Gaussian noises with constant variances $\sigma_{cv,p_x}^2, \sigma_{cv,p_y}^2, \sigma_{cv,v_x}^2, \sigma_{cv,v_y}^2$, respectively. Using~\eqref{eq:cv_model_wrong}, we have:
\begin{equation}\label{eq: P_cv}
    \mathbb{P}(X_t^{cv}| X_{t-1}) = \mathcal{N}(AX_{t-1} , \Sigma_{cv}) := \mathcal{N}(\mu_{cv, t}, \Sigma_{cv}), 
\end{equation}
where $\Sigma_{cv} \in \mathbb{R}^{4\times 4}$ is the diagonal covariance matrix with $[\sigma_{cv,p_x}^2, \sigma_{cv,p_y}^2, \sigma_{cv,v_x}^2, \sigma_{cv,v_y}^2]$ along the diagonal.

In LS-CV, the vehicle's position is projected onto the lane center, and constant velocity is predicted along the lane. In the Frenet-Serret frame~\cite{wang2022research}, let $(s_{t-1}, 0)$ denote the projection of the vehicle position $(p_{x,t-1}, p_{y, t-1})$ (in Cartesian coordinates) onto the lane defined by $\ell$. This projection is expressed as $s_{t-1} = \textsc{PROJECT}(p_{x,t-1}, p_{y, t-1}, \ell)$, where $\textsc{PROJECT}()$ returns the s-coordinate of the Frenet coordinates projection onto the lane $\ell$ defined by lane center waypoints. Let $\mathcal{T}^{x}_{(s,d)}$ and $\mathcal{T}^{y}_{(s,d)}$ denote the transformations that convert Frenet coordinates to $x$ and $y$ Cartesian coordinates, respectively. Let $\| v_{t-1}\|  = \sqrt{v_{x, t-1}^2 + v_{y, t-1}^2}$ be the vehicle speed and $v_s$ be the longitudinal velocity in Frenet frame with magnitude $\| v_{t-1}\|$. We model the LS-CV prediction as:
\begin{align}
X_t^{ls} &= \begin{bmatrix}   
\mathcal{T}^{x}_{(s,d)}(s_{t-1}+\|v_{t-1}\|\Delta t, 0) \\
\mathcal{T}^{y}_{(s,d)}(s_{t-1}+\|v_{t-1}\|\Delta t, 0) \\
\mathcal{T}^{x}_{(s,d)}(v_s,0) \\ 
\mathcal{T}^{y}_{(s,d)}(v_s,0) 
\end{bmatrix} + \begin{bmatrix}
        \epsilon_{p_x}^{ls} \\ \epsilon_{p_y}^{ls} \\ \epsilon_{v_x}^{ls}  \\ \epsilon_{v_y}^{ls}
    \end{bmatrix} \nonumber \\
    &:= g(X_{t-1})+ \epsilon^{ls} \nonumber\\
    &\overset{(1)^*}{\approx} g(\mu_{t-1}) + \nabla g(\mu_{t-1}) (X_{t-1}-\mu_{t-1}) + \epsilon^{ls} \label{eq:ls-cv_linear_wrong},
\end{align}
where $s_{t-1} = \textsc{PROJECT}(p_{x, t-1}, p_{y, t-1}, \ell)$, and $\epsilon_{p_x}^{ls}, \epsilon_{p_y}^{ls}, \epsilon_{v_x}^{ls}, \epsilon_{v_y}^{ls}$ are the zero mean Gaussian noises with constant variances $\sigma_{ls,p_x}^2, \sigma_{ls,p_y}^2, \sigma_{ls,v_x}^2, \sigma_{ls,v_y}^2$, respectively. The approximation $(1)^*$ is driven by the first-order Taylor series expansion of $g(X_{t-1})$ around the mean of the previous state $\mu_{t-1}$, where $\nabla g(\mu_{t-1})$ is the Jacobian matrix containing the gradients of $g(X_{t-1})$ at $\mu_{t-1}$. \begin{figure}
    \centering
    \includegraphics[trim={0 0 0 1cm},clip,width=0.9\columnwidth]{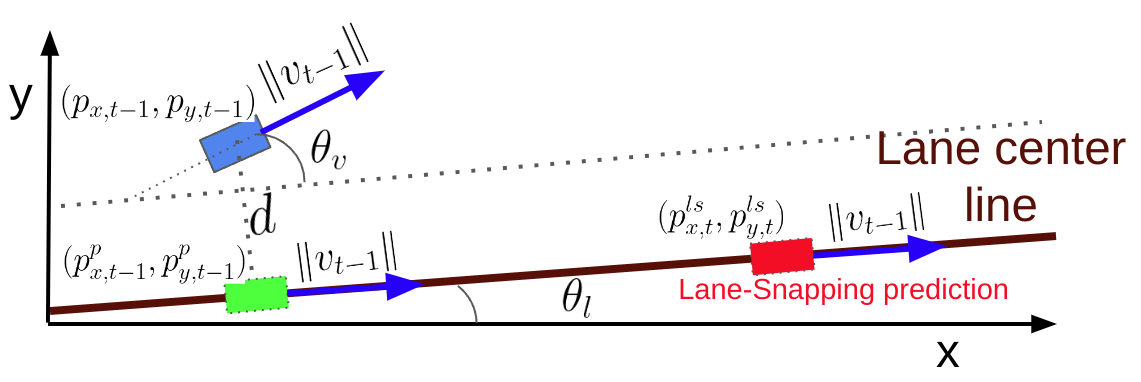}
    \caption{Lane Snapping prediction by locally approximating lane center as a line. The blue vehicle shows the vehicle position and heading at time $t-1$. The green vehicle shows the lane projection of the vehicle position at time $t-1$ and the red vehicle shows the lane snapping prediction at time $t$.}
    \label{fig:jacobian}
\end{figure}

Estimating $\nabla g$ for a general curve is challenging and does not have a closed-form expression due to the Cartesian to Frenet and Frenet to Cartesian transformations. To this end, we locally approximate the lane center curve around the projection point using a line. Let $\theta_{l}$ be the slope of the lane center curve at the point of projection, and hence the slope of the approximate line-center line. Let $d$ be the lateral displacement of the vehicle at point $(p_{x, t-1}, p_{y, t-1})$ w.r.t the lane-center line as shown in Fig.~\ref{fig:jacobian}. Let $\theta_v$ be the vehicle heading w.r.t the lane-center line. In this case, the following suffices:
\begin{align}
    &g(X_{t-1}) = \begin{bmatrix}
        p^{ls}_{x,t} \\ p^{ls}_{y,t} \\  v^{ls}_{x,t} \\
        v^{ls}_{y,t}
    \end{bmatrix}
    \approx \begin{bmatrix}
        p_{x,{t-1}}^p + \|v_{t-1}\|\Delta t \cos \theta_{l} \\
        p_{y,{t-1}}^p + \|v_{t-1}\|\Delta t \sin \theta_{l} \\
        \|v_{t-1}\|\cos \theta_l \\
        \|v_{t-1}\|\sin \theta_l
    \end{bmatrix}, \text{where} \\
    &(p_{x, t-1}^p, p_{y, t-1}^p) = (p_{x, t-1}+d\sin \theta_l, p_{y, t-1}-d\cos \theta_l),    \nonumber \\
    &\|v_{t-1}\| = \sqrt{v_{x,t-1}^2 + v_{y, t-1}^2}. \nonumber
\end{align}

Using geometry, the lateral displacement corresponding to point $(p_{x,t-1} + \Delta x, p_{y,t-1} + \Delta y)$ can be written as $d_{\Delta x, \Delta y}= d-\Delta x \sin \theta_l + \Delta y \cos \theta_l$. Therefore, it is easy to show that $\frac{\partial p^p_{x, t-1}}{\partial p_{x, t-1}} = \cos^2 \theta_l$, $\frac{\partial p^p_{y, t-1}}{\partial p_{y, t-1}} = \sin^2 \theta_l$, and $\frac{\partial p^p_{x, t-1}}{\partial p_{y, t-1}} = \frac{\partial p^p_{y, t-1}}{\partial p_{x, t-1}} = \sin \theta_l \cos \theta_l$. Hence, $\nabla g(X_{t-1})$ can be approximated as:
\begin{equation}\label{eq:nabla}
    \nabla g
    \approx \begin{bmatrix}
        \cos^2 \theta_l & \sin \theta_l \cos \theta_l & \frac{v_x\Delta t \cos \theta_l}{\|v_{t-1}\|} & \frac{v_y\Delta t \cos \theta_l}{\|v_{t-1}\|} \\
        \sin \theta_l \cos \theta_l & \sin^2 \theta_l & \frac{v_x\Delta t \sin \theta_l}{\|v_{t-1}\|} & \frac{v_y\Delta t \sin \theta_l}{\|v_{t-1}\|} \\
        0 & 0 & \frac{v_x \cos \theta_l}{\|v_{t-1}\|} & \frac{v_y \cos \theta_l}{\|v_{t-1}\|} \\
       0 & 0 & \frac{v_x \sin \theta_l}{\|v_{t-1}\|} & \frac{v_y \sin \theta_l}{\|v_{t-1}\|} \\
    \end{bmatrix}, 
\end{equation}
where $\frac{v_x}{\|v_{t-1}\|} = \cos (\theta_l+\theta_v)$ and $\frac{v_y}{\|v_{t-1}\|} = \sin (\theta_l+\theta_v)$.

Using~\eqref{eq:ls-cv_linear_wrong}, we have:
\begin{align}\label{eq:ls-cv_p_wrong}
      \mathbb{P}(X_t^{ls}| X_{t-1}, \ell) &=  \mathcal{N}( \mu_{ls,t} , \Sigma_{ls} ),
\end{align}
where $\mu_{ls,t} := g(\mu_{t-1})+ \nabla g(\mu_{t-1}) (X_{t-1}-\mu_{t-1})$ and $\Sigma_{ls} \in \mathbb{R}^{4\times 4}$ is the diagonal covariance matrix with $[\sigma_{ls,p_x}^2, \sigma_{ls,p_y}^2, \sigma_{ls,v_x}^2, \sigma_{ls,v_y}^2]$ along the diagonal.

\begin{remark}
The constant variances $\sigma_{ls,p_x}^2$, $\sigma_{ls,p_y}^2$, $\sigma_{ls,v_x}^2$, $\sigma_{ls,v_y}^2$ can be modeled as a function of the initial vehicle angular difference and lateral displacement with the lane center. This ensures that the LS-CV model's confidence is higher when the vehicle is laterally closer to a lane with a smaller angular difference.
\end{remark}

Let $I$ denote the identity matrix. Let $K \in \mathbb{R}^{4 \times 4}$ and $\Sigma \in \mathbb{R}^{4 \times 4}$ be diagonal matrices with diagonal entries $K_{i,i} = \frac{\sigma^2_{cv, X_i}}{\sigma^2_{cv, X_i} + \sigma^2_{ls, X_i}}$ and $\Sigma_{i,i} = \frac{\sigma^2_{cv, X_i}\sigma^2_{ls, X_i}}{\sigma^2_{cv, X_i} + \sigma^2_{ls, X_i}}$, respectively, where $\sigma^2_{m, X_1} := \sigma^2_{m, p_x}$, $\sigma^2_{m, X_2} := \sigma^2_{m, p_y}$, $\sigma^2_{m, X_3} := \sigma^2_{m, v_x}$, $\sigma^2_{m, X_4} := \sigma^2_{m, v_x}$, for $m \in \{cv, ls\}$.
    Let the prior belief be given by $\mathbb{P}(p_{t-1}) = \mathcal{N}(\mu_{t-1}, \Sigma_{t-1})$. The GLK method merges two prediction models by exploiting their joint prediction, i.e., $\mathbb{P}(X_t^{GLK} = X_t| \ell):= \mathbb{P}(X_t^{cv} = X_t, X_t^{ls} = X_t| \ell)$. Now we show that the GLK prediction at each timestep is a Gaussian for which we determine the mean and the variance.

\begin{theorem}
    Given the mean $\mu_{t-1}$ and the variance $\Sigma_{t-1}$ of the Gaussian prediction at $t-1$, i.e, $\mathbb{P}(X_{t-1}) = \mathcal{N}(\mu_{t-1}, \Sigma_{t-1})$, the GLK prediction at time $t$ is given by $\mathcal{N}(\mu_{GLK,t}, \Sigma_{GLK,t})$, where $\mu_{GLK,t}=(I-K)A\mu_{t-1} + Kg(\mu_{t-1})$ and $\Sigma_{GLK, t} = M\Sigma_{t-1}M^{\top} + \Sigma$, where $M:=(I-K)A + K \nabla g(\mu_{t-1})$.
\end{theorem}

\begin{proof}
Using~\eqref{eq: P_cv} and~\eqref{eq:ls-cv_p_wrong}, we have:
\begin{align*}
    \mathbb{P}(X_t^{cv}, X_t^{ls}| X_{t-1},\ell) 
    &\overset{(2)^*}{=}  \mathbb{P}(X_t^{cv}| X_{t-1},\ell) \mathbb{P}(X_t^{ls}| X_{t-1},\ell) \\
    &=\mathcal{N}(\mu_{X}, \Sigma_{X}), 
\end{align*}
where equality $(2)^*$ is driven by the independence of two prediction models and $\mu_X = (I-K)\mu_{cv, t} + K\mu_{ls,t} =MX_{t-1} + N$ and $\Sigma_X = \Sigma$, 
where $M:=(I-K)A + K \nabla g(\mu_{t-1})$ and $N:=K(g(\mu_{t-1}) -\nabla g(\mu_{t-1})\mu_{t-1})$. We have:
\begin{align}
        \mathbb{P}(X_t^{GLK}| \ell)  &= \int_{X_{t-1}} \mathbb{P}(X_t^{cv}, X_t^{ls}| X_{t-1},\ell) \mathbb{P}(X_{t-1}| \ell) \nonumber \\
        &=   \int_{X_{t-1}} \mathcal{N}(\mu_{X}, \Sigma_{X}) \mathcal{N}(\mu_{t-1}, \Sigma_{t-1}) \nonumber \\
        &\overset{(3)*}{=} \mathcal{N}(\mu_{GLK, t}, \Sigma_{GLK, t}),
\end{align}
where equality $(3)^*$ is obtained similar to the belief update step in Kalman filter~\cite{welch1995introduction} and  $\mu_{GLK, t}$ and $\Sigma_{GLK, t}$ are given by:
\begin{align}
    \mu_{GLK, t} &= M\mu_{t-1}+N = (I-K)A\mu_{t-1} + Kg(\mu_{t-1}) \\
    \Sigma_{GLK, t} &= M\Sigma_{t-1}M^{\top} + \Sigma_X .
\end{align}
\end{proof}

For brevity and simplicity, we assume that $\sigma_{cv,p_x}^2 = \sigma_{cv,p_y}^2 = \sigma_{cv,v_x}^2 = \sigma_{cv,v_y}^2 := \sigma^2_{cv}$ and $\sigma_{ls,p_x}^2 = \sigma_{ls,p_y}^2 = \sigma_{ls,v_x}^2 = \sigma_{ls,v_y}^2 := \sigma^2_{ls}$ 
and present the base GLK algorithm (Algorithm~\ref{alg:base_wrong}). Interested users are free to consider different noise variances and adapt the algorithm. In Algorithm~\ref{alg:base_wrong}, $V(v_{x, t-1}, v_{y, t-1}, \ell)$ aligns the vehicle speed along the lane center and $\mathcal{T}_{(s,d)^{(x,y)}}$ denotes the conversion from Frenet to Cartesian coordinates.

\begin{algorithm}[hbt!]
\small
\caption{Gaussian Lane Keeping (Base model)}\label{alg:base_wrong}
\textbf{Input: }vehicle position $p_0$, velocity $v_0$, lane center $\ell$, 
time step $\Delta t$, time horizon $T$,
variances $\sigma_{cv}^2$ (constant velocity) and $\sigma_{\ell}^2$ (lane snapping), \\
\textbf{Output: }{prediction vector $P$}\\
\textbf{Initialize: } $t = 1$, $P = []$, Identity matrix $I \in \mathbb{R}^{4 \times 4}$, Matrix $A$ (see \eqref{eq:cv_model_wrong}), $X_{0} = [p_{x,0}, p_{y,0}, v_{x,0}, v_{y,0}]^{\top}$ \\
$\mu_{t-1} \gets X_0$, $\Sigma_{t-1} \gets 0$ \ \ \blue{Prior mean and Variance} \\
$K \gets \frac{\sigma_{cv}^2}{(\sigma_{cv}^2 + \sigma^2_{\ell})}I$ \ \ \blue{Compute $k$}\\
$\Sigma \gets \frac{\sigma_{cv}^2\sigma_{ls}^2}{(\sigma_{cv}^2 + \sigma^2_{\ell})}I$ \ \ \blue{Compute $\Sigma$}\\
\For{$t \le T$}{
    
    $\mu_{cv,t} \gets A\mu_{t-1}$ \ \ 
    \blue{Constant velocity prediction}\\
     $(p_{x,t-1}, p_{y,t-1}, v_{x,t-1}, v_{y,t-1})
     \gets \mu_{t-1}$ \ \blue{Prior estimate} 
     \\
    $(s_{t-1}, 0) \gets \textsc{PROJECT}(p_{x, t-1}, p_{y, t-1} \ell)$ \ \  \blue{Projection on $\ell$ in Frenet frame} \\
    $s_t \gets s_{t-1}+\|v_{t-1}\|\Delta t$ \ \ \blue{Lane snapping position prediction in Frenet frame} \\
    $v_{s,t} \gets  \textsc{V}(v_{x, t-1}, v_{y, t-1}, \ell)$ \blue{Lane snapping velocity prediction in Frenet frame} \\
    $\mu_{ls, t} \gets \mathcal{T}^{(x,y)}_{(s,d)}(s_{t}, 0), \mathcal{T}^{(x,y)}_{(s,d)}(v_{s, t}, 0)$ \ \ \blue{Calculate lane snapping prediction by converting to Cartesian frame} \\
    $\mu_{GLK, t} \gets (I-K)\mu_{cv,t} + K  \mu_{ls,t}$ \  \blue{GLK mean}\\

Estimate $\nabla g(\mu_{t-1})$ using \eqref{eq:nabla} \ \ \ \blue{Estimate Jacobian} \\
$M \gets (I-K)A + K\nabla g(\mu_{t-1}) $  \ \ \blue{Calculate $M$}\\    
    $\Sigma_{GLK,t} \gets M\Sigma_{t-1}M^{\top} + \Sigma$ \ \ \ \blue{GLK variance}\\
    $P \gets \textsc{append}(P, \mu_{GLK, t}, \Sigma_{GLK,t})$   \ \ \ \  \blue{Append $\mu_{GLK,t}$ and $\Sigma_{GLK, t}$ to prediction vector $P$}\\
    $\mu_{t-1} \gets \mu_{GLK, t}$, $\Sigma_{t-1} \gets \Sigma_{GLK, t}$ \ \ \blue{Update prior}
}
\end{algorithm}

\normalsize


\begin{figure}[ht]
    \centering
    \includegraphics[trim={0 0 0 1cm}, clip,width=0.9\columnwidth]{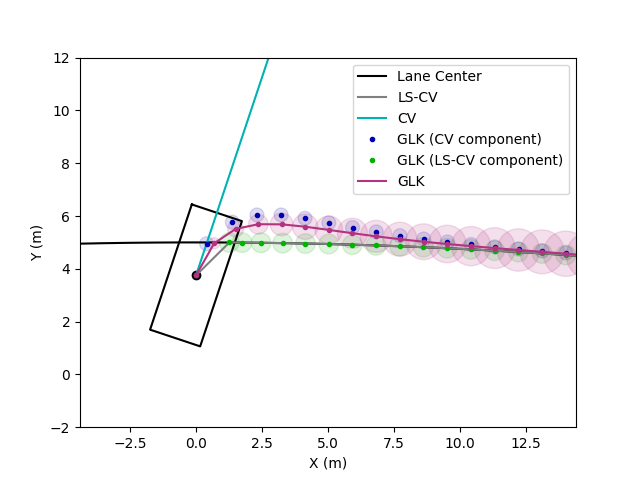}
    \caption{Visualization of GLK. CV predictions (cyan) indicate the car moving forward with a constant heading. LS-CV model (gray) assumes the car stays in the lane. The GLK prediction (magenta) shows the resulting prediction formed from the combination of CV (dark blue) and LS-CV (green) components at each timestep.}
    \label{fig:glk_iterative}
\end{figure}

Figure~\ref{fig:glk_iterative} provides an illustrative example for GLK compared to other baselines. Although the figure shows a significant angular difference between the vehicle heading and the lane, this is primarily for demonstrative purposes. In situations of extreme angular differences (e.g., surpassing a threshold such as $\pi/6$), it is often reasonable to consider the lane-keeping assumption invalid. Consequently, transitioning exclusively to a constant velocity model, or a curvature variant as detailed in Sec.~\ref{sec:extentions}, may be more appropriate.


\section{Extensions}\label{sec:extentions}

\subsection{Curvature}
In scenarios involving vehicular turns, i.e., $\theta_{t} - \theta_{t-1} > \epsilon$, where $\theta_{t}$ is the heading angle at time $t$ and $\epsilon$ is a small angle, the constant heading assumption in the constant velocity model can be relaxed by the inclusion of the rate of curvature. Let $\Delta\theta_{t,t-1} = \theta_{t} - \theta_{t-1}$ denote the change in vehicle heading.  Since a direct application: $\theta_{t+1} \leftarrow \theta_{t} + \Delta\theta_{t,t-1}$ will overly prolong turns, the vehicle heading can be modeled such that the turning decays over the prediction horizon:
\begin{align}
\theta_{t+i+1} \leftarrow \theta_{t+i} + d^i \delta\theta_{t+i,t+i-1},
\end{align}
where $i$ indexes the prediction step and the rate of decay $d$ is $0\le d \le 1$. Lower values of $d$ results in a faster decay.

\subsection{Multi-modal Predictions}

The trajectory of vehicles is often dependent on the agent's intentions. Without insight into these intentions, distinguishing among several potential diverging trajectories becomes challenging. Hence, it's often critical to identify multiple most likely modes to precisely anticipate vehicle behavior. Multi-modal predictions are an active area of research \cite{cui2019multimodal,li2020socially}, and it is desirable to have robust baselines to benchmark the performance of these multi-modal prediction methods.


To infuse our baseline models with multi-modal capabilities, we employ a method where a vehicle can be linked with multiple lane centers based on predefined lateral distance and orientation thresholds concerning each lane center. When a vehicle is associated with multiple lane centers, the prediction model generates trajectory predictions corresponding to each of these associated lane centers, thus producing a set of multi-modal predictions. This approach allows us to harness the diverse possibilities inherent in vehicle motion and ensures that our predictions are not overly constrained, ultimately leading to more accurate and robust results. In the case of multi-modal prediction, probabilities for each predicted trajectory can be estimated based on the heading angle, or learned to model the likelihood of each trajectory.  


\subsection{Interactive Predictions}

One significant limitation of the GLK model with constant velocity lies in its lack of interactivity. In this configuration, predictions for individual vehicles remain isolated, overlooking any potential inter-agent interactions. Predicting interactive behaviors is an active area of research \cite{10588713}, and to facilitate meaningful progress, it is essential to have baseline prediction algorithms pertinent to the studied phenomena. To augment the interactivity of the GLK model without substantially compromising its robustness, we incorporate the Intelligent Driver Model (IDM) (see Section~\ref{subsec:IDM}) to determine the velocity of each vehicle.



In our approach, we integrate the IDM model to calculate the velocity of each vehicle but selectively apply it only when there is a leading vehicle ahead. When no leading vehicle is present, we maintain a constant velocity for that particular vehicle. Employing constant velocity in the absence of a leading vehicle offers a crucial performance advantage. This is because the IDM model assumes that all stationary cars with no leading vehicle will accelerate. However, in reality, a stationary car with no vehicle in front of it often remains stationary, making a constant velocity prediction the more accurate choice in that scenario.


 
Recall from \eqref{eq:IDM_a} and \eqref{eq:IDM_s}, that IDM model necessitates IDM parameters $v_0$, $s_0$, $s_1$, $T$, $a_{\max}$, $b$.
We employ a particle filter~\cite{thrun2002particle} for estimating the IDM parameters online for each vehicle. At any instance, the IDM parameters associated with the particle possessing the highest weight are employed to compute the vehicle's acceleration. This, in turn, enables us to predict the vehicle's velocity at the subsequent time step, using $v_{t+1} = v_t + a \Delta t$.


\section{Metrics}\label{sec:Metrics}

Average Displacement Error (ADE) and Final Displacement Error (FDE) are two widely used metrics for measuring prediction accuracy. Additionally, specialized metrics tailored for specific functionalities, such as overlap rate\footnote{https://waymo.com/open/challenges/2023/motion-prediction} and cross-track errors~\cite{cui2019multimodal}, have been proposed in the literature. In the case of multi-modal predictions, the prediction exhibiting the lowest ADE and FDE concerning the ground truth is typically considered. For example, in a scenario where a prediction model generates two trajectories for a vehicle approaching an intersection, one for a left turn and the other for moving straight ahead, the ground truth may align closely with only one of these possibilities contingent upon the driver's intent. In such a case, the prediction closest to the ground truth is used for the evaluation.

When using ADE as a performance benchmark, many learning-based approaches gain an advantage by directly optimizing ADE through its use as a loss function. However, it's crucial to acknowledge that ADE and FDE primarily prioritize accuracy in prevalent driving scenarios, often averaging out numerous subpar predictions in the less frequent edge cases. As our community progressively gravitates towards more sophisticated prediction methodologies to address the long tail of outliers, relying solely on ADE and FDE tends to obscure the meaningful advancements we truly seek.

In our efforts to pinpoint and refine predictions, especially in outlier situations, we have found it beneficial to analyze a plot of sorted errors. Viewing all prediction errors (instead of a single average) visually provides deeper insights into understanding an algorithm's capabilities. Alternatively, we can select a single algorithm to establish the sorting order, enabling more direct comparisons of strengths and weaknesses between algorithms. This comparison method simplifies visualization and debugging of failures.

However, it's important to note that this method doesn't always align error size with impact. For example, a car crossing into our lane might exhibit a relatively minor lateral error, yet the safety implications could be significant.

\section {Evaluation}\label{Sec: Evaluation}

We now discuss the evaluation of the prediction models. 

\subsection {Comparison with Deep Learning Method}\label{Sec:dl}
As mentioned in the background section, deep learning methods often do not transfer well to new datasets. To illustrate this, we consider GATraj~\cite{cheng2023gatraj}, a sophisticated model for multi-agent trajectory prediction that leverages graph and attention mechanisms. GATraj utilizes a graph convolutional network to capture intricate inter-agent interactions. It has demonstrated state-of-the-art prediction performance on benchmark datasets such as ETH~\cite{pellegrini2009you}/UCY~\cite{leal2014learning} for pedestrian trajectories and nuScenes dataset~\cite{caesar2020nuscenes} for autonomous driving~\cite{cheng2023gatraj}. Our selection of GATraj is motivated by the availability of its open-source implementation and a pre-trained model, which facilitates our evaluation of its capabilities on a challenging out-of-distribution dataset. 

In contrast, many other neural-network-based prediction models~\cite{varadarajan2022multipath++, luo2023jfp} that claim to achieve state-of-the-art results present significant barriers to entry. These models often lack open-source implementations, necessitate time-consuming model re-training, or demand extensive implementation efforts for adaptation to datasets beyond their original training data. These constraints severely limit their practical utility beyond their specific domains and impede their suitability as dependable baselines for comparison. 

\begin{figure}
    \centering \includegraphics[width=0.7\columnwidth]{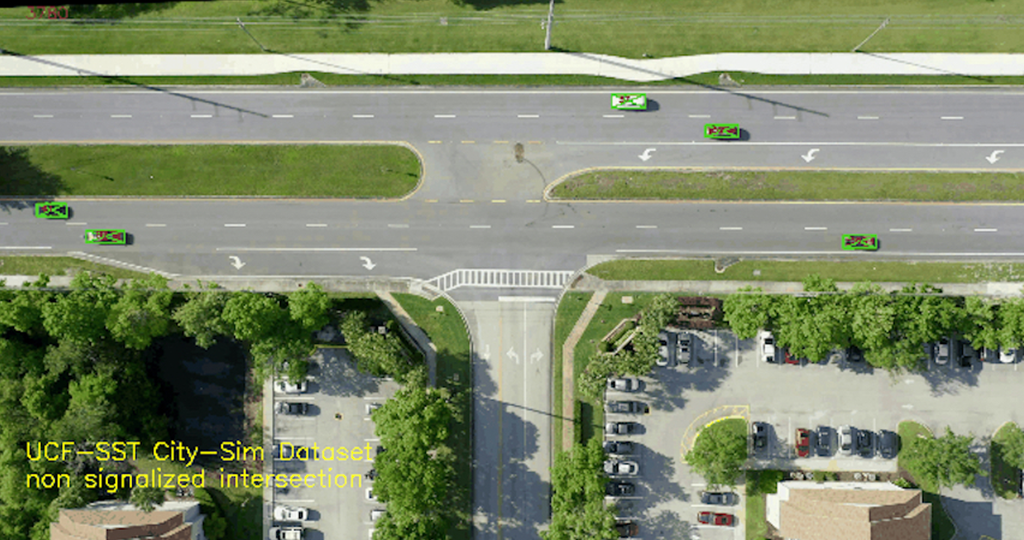}
    \caption{Intersection B scenario from the CitySim Dataset}
    \label{fig:intersectionB}
\end{figure}

We conducted a performance assessment of GATraj on the Intersection B scenario of the CitySim Dataset\footnote{https://github.com/ozheng1993/UCF-SST-CitySim-Dataset}, as depicted in Fig.~\ref{fig:intersectionB}. This scenario encompasses CSV files containing agent trajectories recorded at different times. For our evaluation, we focused on the first $100$ seconds of data from the first two trajectory files, denoted as Intersection-B-01 and Intersection-B-02 onward.

The GATraj model utilizes each agent's historical data for the past $8$ time steps and predicts their future trajectories for the subsequent $12$ time steps. We compute the ADE and FDE for a prediction horizon of $6$ seconds by filtering the Intersection B scenario at intervals of $0.5$ seconds. Detailed results for ADE and FDE can be found in Table~\ref{tab:evaluation_without_ignore}.

\begin{figure}
 \centering
 \begin{subfigure}[b]{0.2\textwidth}
\includegraphics[width=1\linewidth, height=1\linewidth, keepaspectratio]{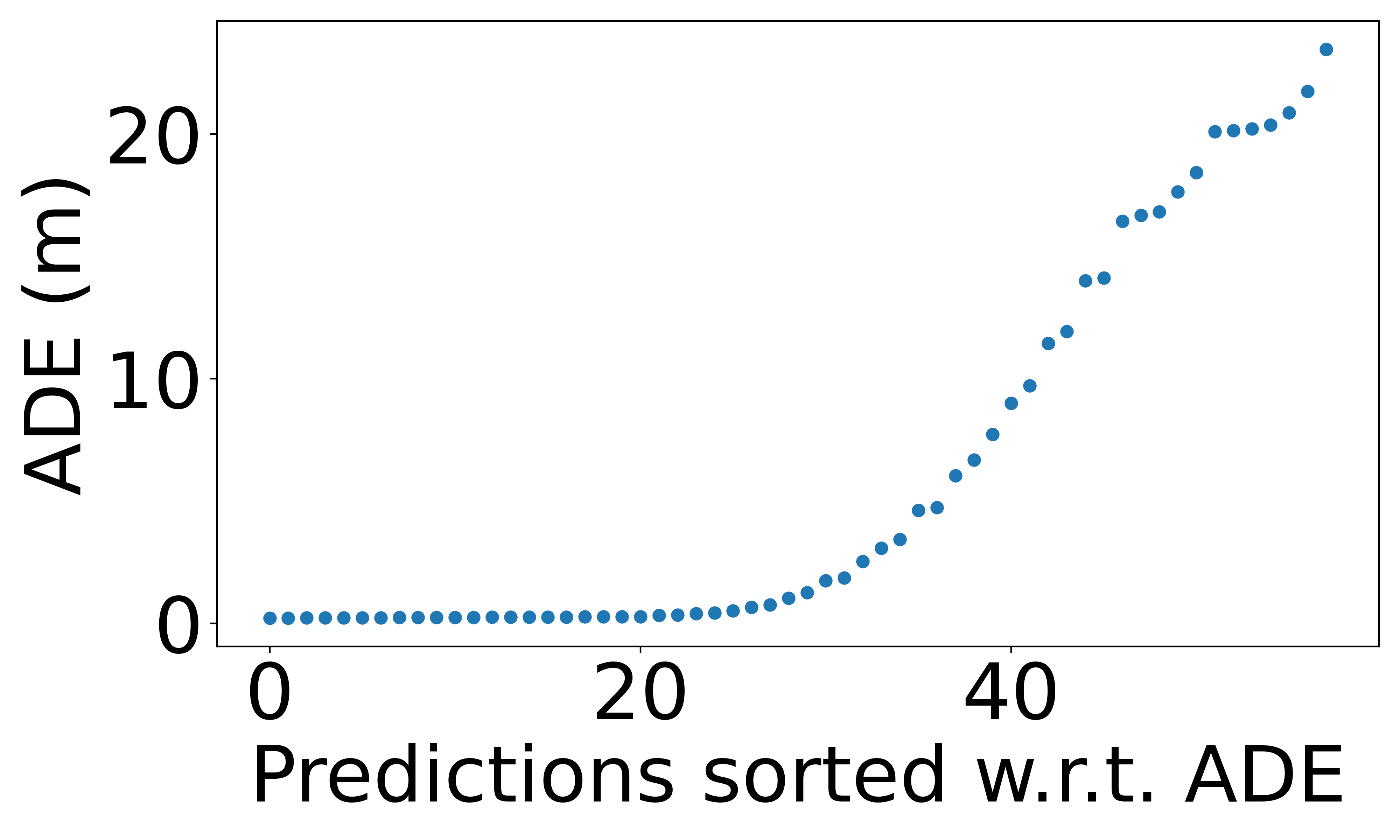}
 \caption{}
 \label{fig:ade_1}
 \end{subfigure}
~~
 \begin{subfigure}[b]{0.2\textwidth}
 \centering
\includegraphics[width=1\linewidth, height=1\linewidth, keepaspectratio]{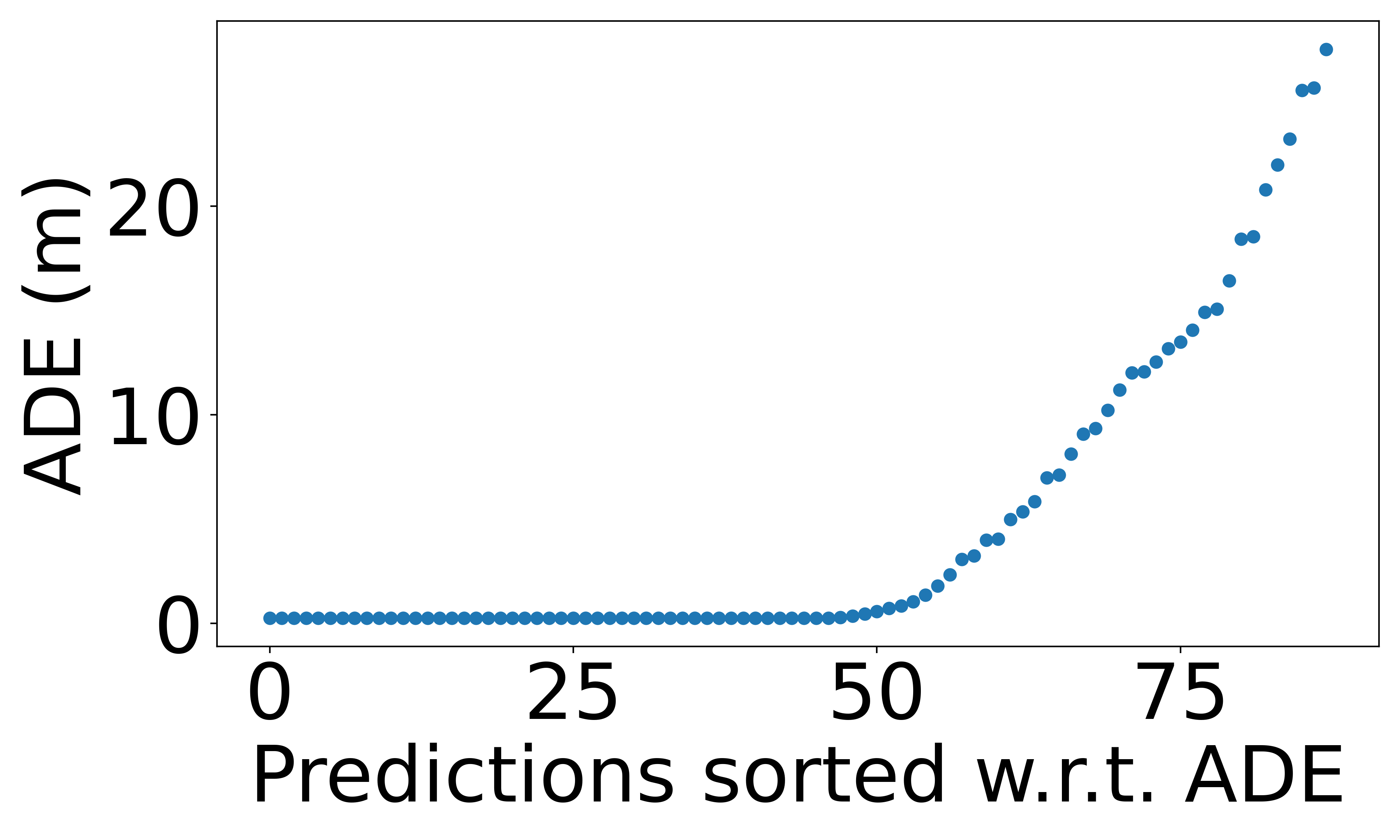}
 \caption{}  
 \label{fig:ade_2}
 \end{subfigure}

 \begin{subfigure}[b]{0.2\textwidth}
	 \centering	 \includegraphics[width=1.0\linewidth, height=1.0\linewidth, keepaspectratio]{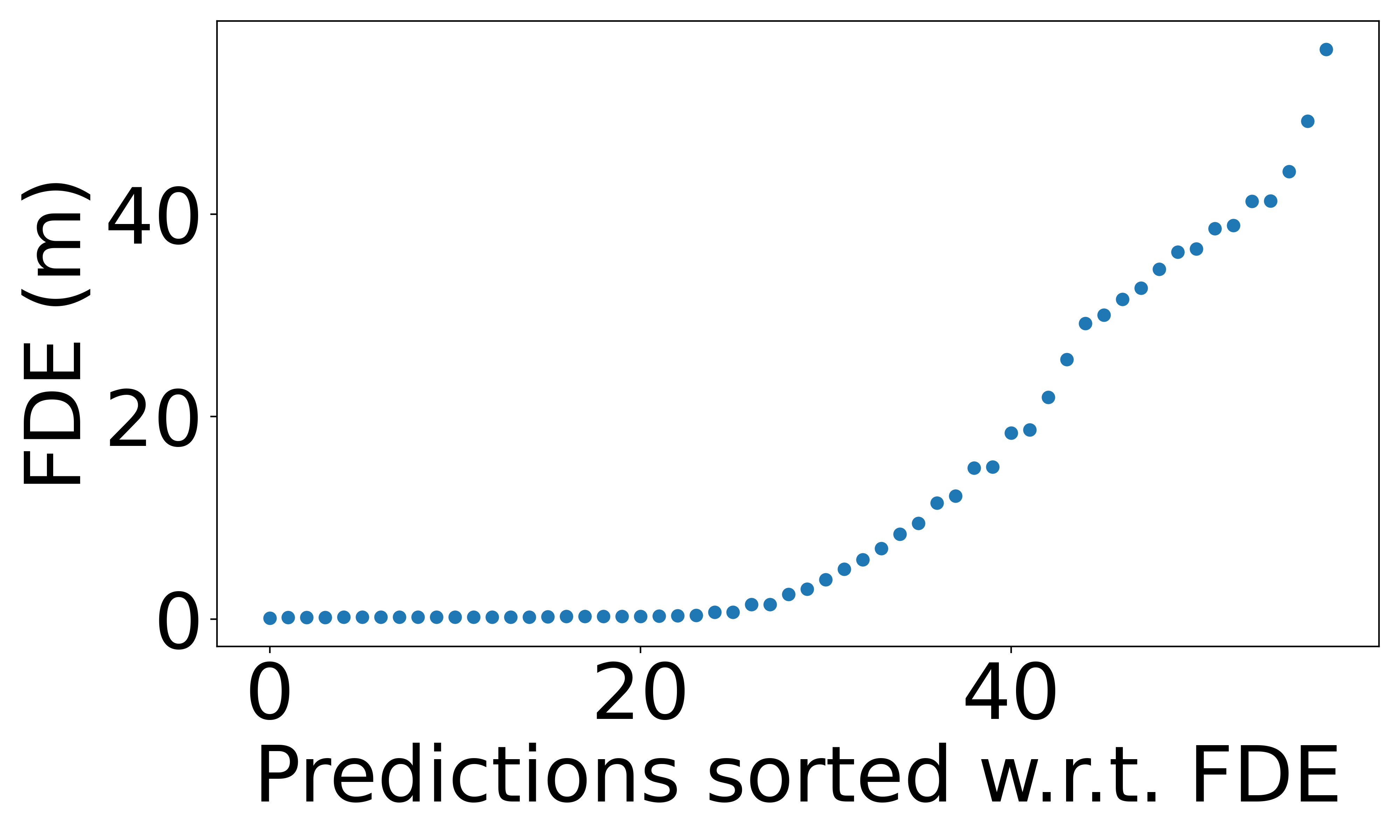}
   \caption{}
 \label{fig:fde_1}
 \end{subfigure}
~~
\begin{subfigure}[b]{0.2\textwidth}
	 \centering
\includegraphics[width=1.0\linewidth, height=1.0\linewidth, keepaspectratio]{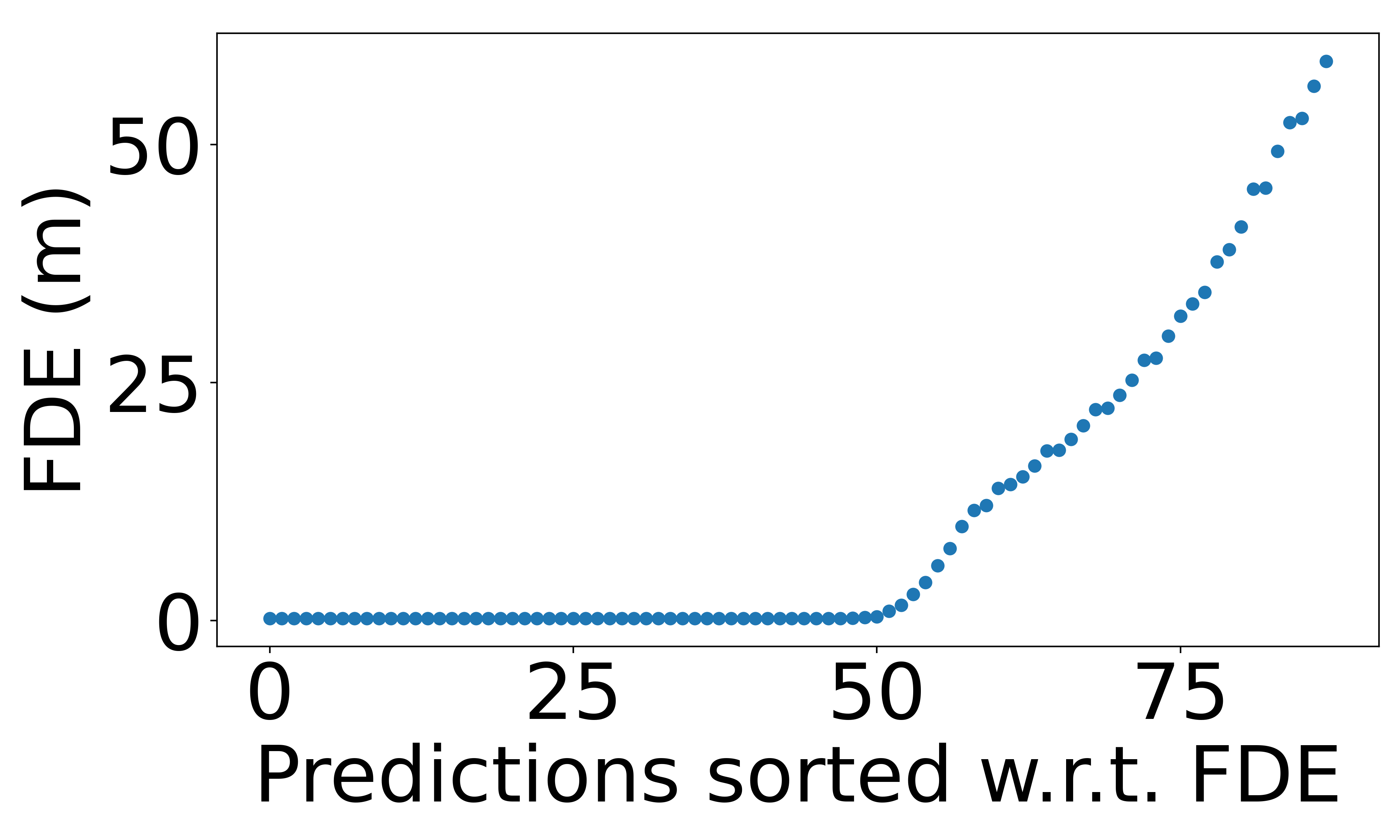}
 \caption{}
 \label{fig:fde_2}
 \end{subfigure}
   \caption{\footnotesize Sorted Errors for GATraj on CitySim (out of distribution) }
\end{figure}

Figure~\ref{fig:ade_1} and \ref{fig:ade_2} showcase the sorted ADE results for Intersection-B-01 and Intersection-B-02, respectively. Correspondingly, Fig.~\ref{fig:fde_1} and \ref{fig:fde_2} showcase sorted FDE results for the same scenarios. 
It's noteworthy that at the higher end of the spectrum, both ADE and FDE exhibit notably large values, which are roughly double the error of the baseline methods -- we detail it in the next section.


\subsection {Baseline Model Evaluations}\label{sec:baseline_eval}

We evaluated five baseline models on the vehicle trajectories from Intersection-B-01 and Intersection-B-02. These models include Constant Velocity (CV), Lane Snapping with Constant Velocity (LS-CV), GLK with Constant Velocity (GLK-CV), Lane Snapping with IDM Velocity (LS-IDM), and GLK with IDM Velocity (GLK-IDM). In Table~\ref{tab:evaluation_without_ignore} reports the ADE and FDE results of the baseline models for a prediction horizon spanning $6$ seconds. This longer prediction horizon of $6$ seconds is chosen based on the trajectory lengths in the CitySim dataset, surpassing the typical $2$-$3$ sec prediction horizon utilized by deep-learning prediction models.
Such extended horizons often pose challenges for prediction, thereby highlighting the robustness of the algorithms. Notably, the baseline models consistently outperform GATraj, a neural network-based prediction model, when applied to the same trajectory data. This exemplifies simple solutions that can outperform highly advanced solutions.

\begin{table}
\centering
\setlength{\tabcolsep}{8pt}
\renewcommand{\arraystretch}{1.2}
\begin{tabular}{|l|c|c|c|c|}
\hline
\textbf{Trajectory Data} & \multicolumn{2}{c|}{\textbf{Intersection-B-01}} & \multicolumn{2}{c|}{\textbf{Intersection-B-02}} \\
\cline{2-5}
\textbf{Model}           & \textbf{ADE}           & \textbf{FDE}           & \textbf{ADE}           & \textbf{FDE}           \\
\hline
CV           & 2.98         &7.75     & 2.60  &  6.91 \\
\hline
LS-CV        & 2.42         & 5.90          & 2.38         & 6.19         \\
\hline
GLK-CV           & \textbf{2.26}         & \textbf{5.47}     & \textbf{2.33}         & \textbf{5.93}         \\
\hline
LS-IDM       & 2.99         & 7.07      & 2.69        & 6.68         \\
\hline
GLK-IDM    & 2.84         & 6.54       & 2.69         & 6.39         \\
\hline
\hline
GATraj    & 6.15         & 12.64       & 4.78         & 10.90         \\
\hline
\end{tabular}
\caption{ADE and FDE for the baseline models for a prediction horizon of $6$ seconds.}
\label{tab:evaluation_without_ignore}
\end{table}

\begin{figure}[t]
    \centering
    \includegraphics[trim={0 0 0 0cm},clip,width=0.9\columnwidth]{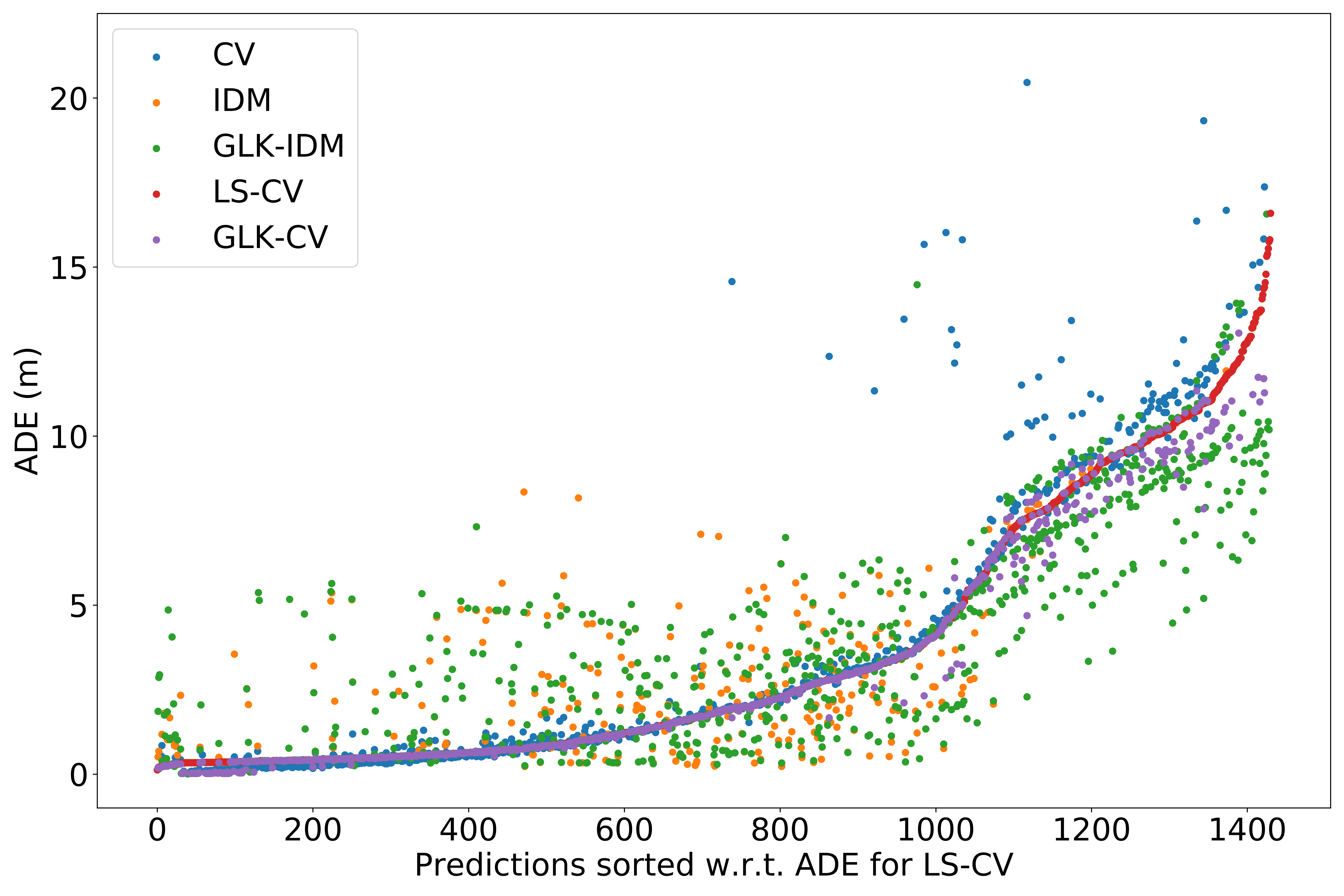}
    \caption{Predictions of the baseline models sorted w.r.t the ADE of the lane snapping with constant velocity model. 
    }
    \label{fig:sorted_predictions_0_ignore}
\end{figure}

In Fig.~\ref{fig:sorted_predictions_0_ignore}, we present predictions from our baseline models, sorted based on the ADE with respect to the LS-CV model. It is evident that IDM-based models exhibit less favorable predictions on the lower end, while generally outperforming the constant velocity-based models (as indicated by their lower position in the chart) at the high end of the error spectrum. This discrepancy can largely be attributed to the non-converged particle filter at the beginning of trajectories, resulting in poorly tuned IDM parameters \cite{moradipari2022predicting}.

\begin{figure}[t]
    \centering
    \includegraphics[width=0.9\columnwidth]{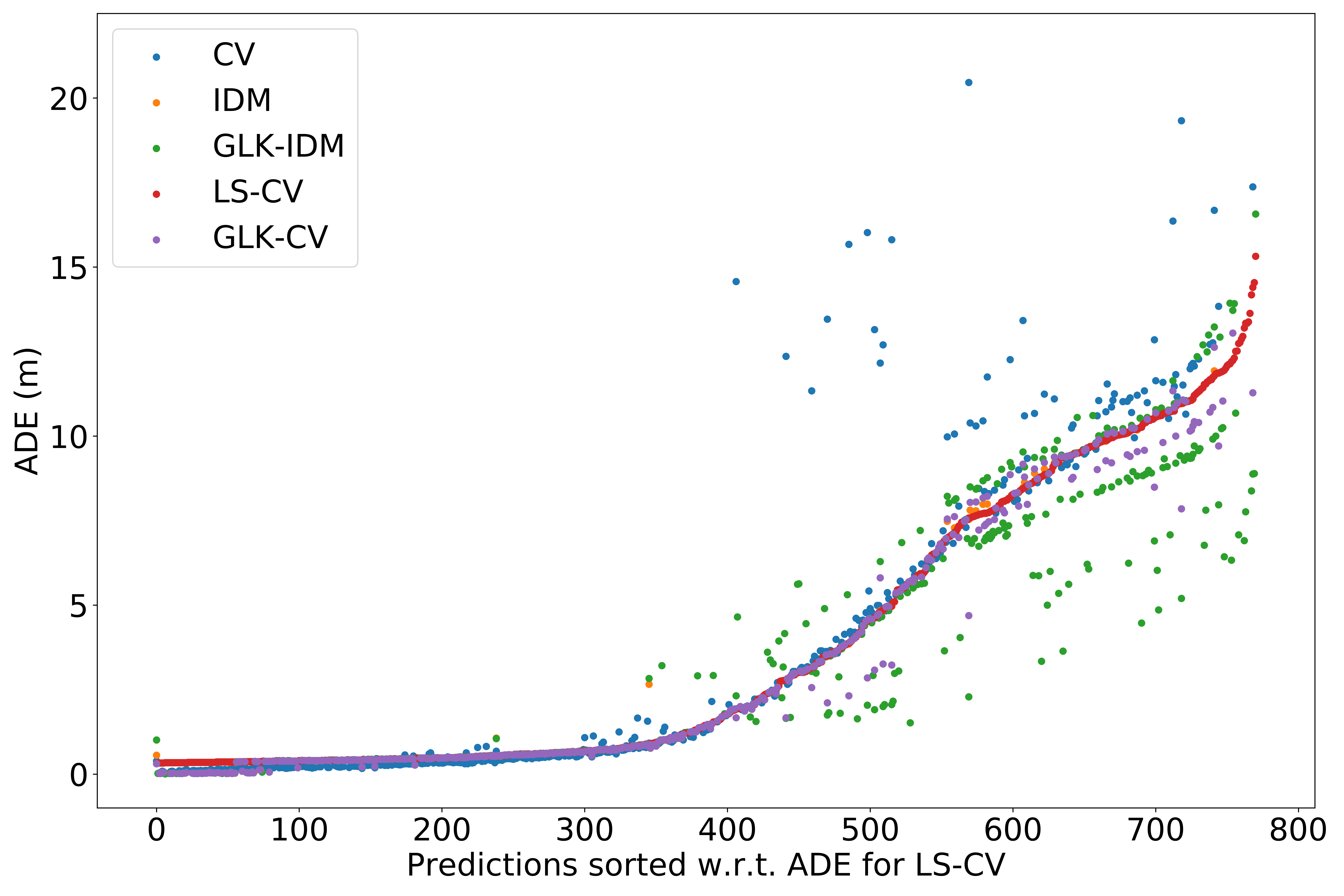}
    \caption{Predictions of the baseline models sorted w.r.t the ADE of the lane snapping with constant velocity model. The predictions made in the first $2$ seconds of when each vehicle first appears in the dataset are ignored.}
    \label{fig:sorted_predictions_2_ignore}
\end{figure}
\begin{table}[ht!]
\centering
\setlength{\tabcolsep}{8pt}
\renewcommand{\arraystretch}{1.2}
\begin{tabular}{|l|c|c|c|c|}
\hline
\textbf{Trajectory Data} & \multicolumn{2}{c|}{\textbf{Intersection-B-01}} & \multicolumn{2}{c|}{\textbf{Intersection-B-02}} \\
\cline{2-5}
\textbf{Model}           & \textbf{ADE}           & \textbf{FDE}           & \textbf{ADE}           & \textbf{FDE}           \\
\hline
CV       &8.30   &23.74  & 9.54  & 28.92 \\
\hline
LS-CV & 4.39 & 10.61 & 6.71 & 17.90 \\
\hline
GLK-CV  & 3.89 & 9.56 & 6.30 & 16.22 \\
\hline
LS-IDM    & 4.57 & 11.05 & 7.19 & 18.86 \\
\hline
GLK-IDM   & \textbf{3.78} & \textbf{9.17} & \textbf{5.13} & \textbf{12.03} \\
\hline
\end{tabular}
\caption{ADE and FDE for the baseline models for a prediction horizon of $6$ seconds by excluding the predictions made in the first $2$ seconds of when each vehicle first appears in the dataset -- the first $2$ seconds are the processing time for the particle filter to converge.}
\label{tab: evaluation_with_2_sec_ignore}
\end{table}
To confirm this hypothesis, we adopt a strategy wherein we allow a grace period for the particle filter to converge and provide more accurate IDM parameter estimates. Consequently, we disregard predictions from all models for the first $2$ seconds of a vehicle's appearance in the dataset. In practice, the constant velocity counterpart of these models can be employed during this period required for the particle filter convergence.

Table~\ref{tab: evaluation_with_2_sec_ignore} reports the ADE and FDE results of the baseline models for a prediction horizon of $6$ seconds, excluding the time for the particle filter to converge. It is worth noting that this approach significantly enhances the performance of IDM-based methods in comparison to the constant velocity-based baselines (see Fig.~\ref{fig:sorted_predictions_2_ignore}).

However, it is essential to acknowledge the larger values of ADE and FDE in Table~\ref{tab: evaluation_with_2_sec_ignore}, as compared to Table~\ref{tab:evaluation_without_ignore}. We hypothesize that this increase in error may be attributed to the road geometry, particularly since the new vehicles initially appear on the straight road segments within the dataset (refer to Fig.~\ref{fig:intersectionB}). Consequently, the initial two seconds correspond to trajectories on relatively straight paths, which are typically easier to predict. These segments, however, are excluded from the analysis in Table~\ref{tab: evaluation_with_2_sec_ignore}.


In conclusion, 
Fig.~\ref{fig:sorted_predictions_2_ignore} reveals that the GLK model with IDM  outperforms the other models, particularly on the higher end of the error spectrum. GLK-CV shows improvements in the low error regime, whereas CV exhibits relatively poorer performance, especially in challenging high-error circumstances.

\begin{figure}[ht]
    \centering
    \includegraphics[width=0.85\columnwidth]{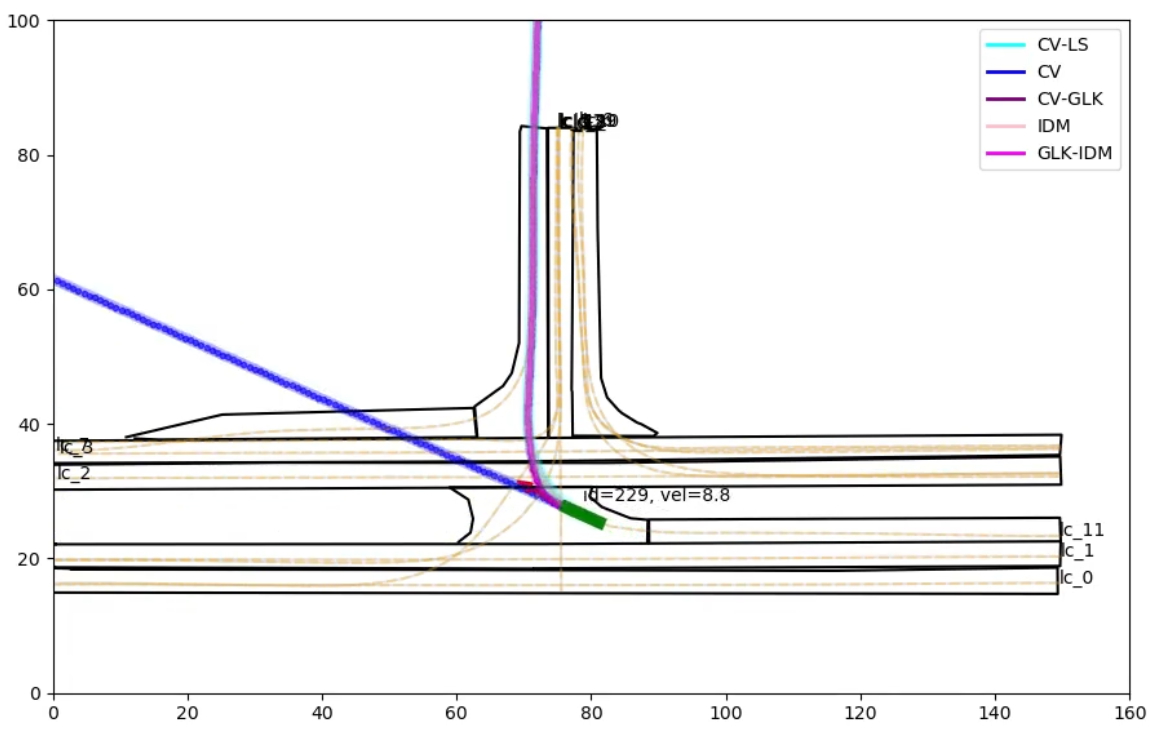}
    \caption{Comparison of case where prediction errors vary significantly.}
    \label{fig:compare}
\end{figure}

\subsection{Qualitative analysis}
We observe comparable results in ADE and FDE from CLK-CV and LS-CV. This is anticipated since GLK is tailored to enhance performance in scenarios where a vehicle deviates from its lane, a relatively infrequent event in typical datasets.

In Fig.~\ref{fig:compare}, we present an instance where GLK outperforms LS. Here, the enhancement can be attributed to a lateral shift that adjusts the curvature. Additionally, it becomes apparent that CV exhibits significant error on curved trajectories, a trend aligned with our expectations.


\subsection{Real Car Demonstration}

The GLK-CV prediction algorithm was utilized in a public vehicle demonstration on an urban city course held in Joso City, Japan, as part of the IFAC Conference in July 2023. This demonstration encompassed various driving scenarios, such as navigating a 4-way intersection, T-junction, and executing lane changes. The inclusion of the prediction algorithm underscored its reliability and applicability for real-world scenarios, marking a significant milestone in showcasing its effectiveness in practical applications.


The test drivers were given scripted behaviors to follow, unknown to the ego vehicle. These scripts were carefully crafted to present challenges to the ego vehicle, including scenarios where oncoming traffic refused to yield, abrupt lane changes by other vehicles, and deliberate acceleration to block gaps targeted by the ego vehicle. These scripted challenges were designed to rigorously test the capabilities of the ego vehicle's decision-making and response mechanisms in complex and dynamic traffic environments.


\begin{figure}[ht]
    \centering
    \includegraphics[height=0.4\columnwidth]{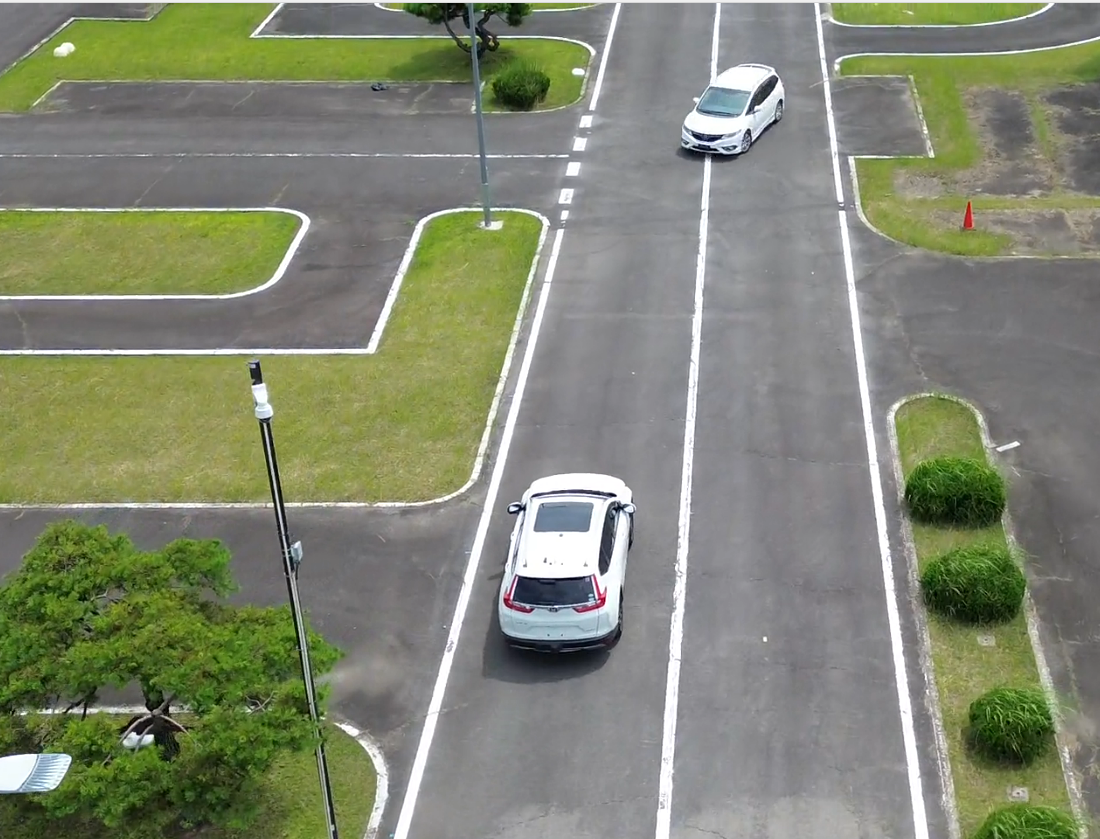}
    \includegraphics[height=0.4\columnwidth]{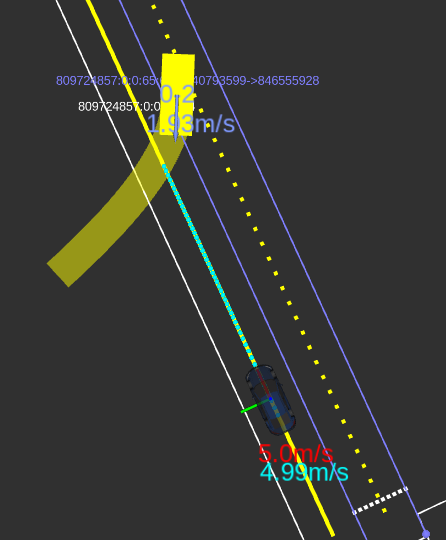}
    \caption{A traffic vehicle turns onto an unmapped road, and GLK-CV is able to accurately predict the crossing behavior.
    }
    \label{fig:rviz}
\end{figure}

The GLK-CV prediction played a crucial role in a specific segment of the course, notably when a traffic vehicle turned onto an unmapped road, as depicted in Fig.~\ref{fig:rviz}. In this scenario, LS-CV predictions would have overlooked the lane departure, potentially leading to a collision. Conversely, the CV prediction would have anticipated a trajectory directly intersecting with the path of the ego car, resulting in an unnecessary hard brake. Feedback from the demo participants highlighted that they felt the scenarios were challenging and that the vehicle responded to changes in a way that felt comfortable. 

\subsection{Summary of Evaluations}
In summary, the validation studies demonstrate the high efficacy of the proposed GLK method. It was evaluated on the CitySim Dataset, outperforming transferred state-of-the-art neural networks (GATraj). 
We expect this analysis to serve as a formal reference for how rule-based methods can be more generalizable than data-driven methods like neural networks. We also expect the presented GLK method to help facilitate autonomous vehicle research as an effective and easily implementable baseline. 
\section{Conclusions}\label{sec:Conclusions}
We present a robust prediction baseline algorithm, Gaussian Lane Keeping (GLK), that combines vehicle heading with lane information. We extend this model to include both interactive and multi-modal predictions. We evaluate the prediction model on the CitySim dataset and utilize it in a public vehicle demonstration in Joso City, Japan. We show how these algorithms improve over both a state-of-the-art deep learning model and constant velocity baseline while preserving robustness.

\bibliographystyle{IEEEtran}
\bibliography{ref}

\end{document}